\documentclass{ws-ijmpc}

\begin{document}

\markboth{Jian-Guo Liu et. al.} {Improved Collaborative Filtering
Algorithm via Information Transformation}

\catchline{}{}{}{}{}
%%%%%%%%%%%%%%%%%%%%%%%%%%%%%%%%%%%%%%%%%%%%%%%%%%%%%%%%%%%%%%%%%%%%

\title{Improved Collaborative Filtering Algorithm via Information Transformation}
%\footnote{For the title, try not to use more than 3 lines.
%Typeset the title in 10 pt roman, uppercase and boldface.} }

\author{JIAN-GUO LIU, BING-HONG WANG}

\address{Department of Modern Physics and Nonlinear Science
Center, University of Science and Technology of China, Hefei 230026,
P R China \\ Department of Physics, University of Fribourg, Chemin
du Mus\'{e}e 3, CH-1700 Fribourg, Switzerland%\footnote{State completely without abbreviations, the
%affiliation and mailing address, including country. Typeset in 8 pt
%italic.}\\
\\
liujg004@ustc.edu.cn, bhwang@ustc.edu.cn}

\author{QIANG GUO}

\address{Dalian Nationalities University, Dalian 116600, P R China\\
guoqiang@dlnu.edu.cn}

\maketitle

\begin{history}
\received{Day Month Year}
\revised{Day Month Year}
\end{history}

\begin{abstract}
In this paper, we propose a spreading activation approach for
collaborative filtering (SA-CF). By using the opinion spreading
process, the similarity between any users can be obtained. The
algorithm has remarkably higher accuracy than the standard
collaborative filtering (CF) using Pearson correlation. Furthermore,
we introduce a free parameter $\beta$ to regulate the contributions
of objects to user-user correlations. The numerical results indicate
that decreasing the influence of popular objects can further improve
the algorithmic accuracy and personality. We argue that a better
algorithm should simultaneously require less computation and
generate higher accuracy. Accordingly, we further propose an
algorithm involving only the top-$N$ similar neighbors for each
target user, which has both less computational complexity and higher
algorithmic accuracy.

\end{abstract}

\ccode{PACS Nos.: 89.75.Hc, 87.23.Ge, 05.70.Ln}

\section{Introduction}
With the advent of the Internet, the exponential growth of the
World-Wide-Web and routers confront people with an information
overload \cite{Broder2000}. We are facing too much data to be able
to effectively filter out the pieces of information that are most
appropriate for us. A promising way is to provide personal
recommendations to filter out the information. Recommendation
systems use the opinions of users to help them more effectively
identify content of interest from a potentially overwhelming set of
choices \cite{Resnkck1997}. Motivated by the practical significance
to the e-commerce and society, various kinds of algorithms have been
proposed, such as correlation-based methods
\cite{Herlocker2004,Konstan1997}, content-based methods
\cite{Balab97,Pazzani99}, the spectral analysis
\cite{Billsus1998,Sarwar2000a}, principle component analysis
\cite{Goldberg2001}, network-based methods
\cite{Zhang2007a,Zhang2007b,Zhou2007,Zhou2007b}, and so on. For a
review of current progress, see Ref. \cite{Adomavicius2005} and the
references therein.

One of the most successful technologies for recommendation systems,
called \emph{collaborative filtering} (CF), has been developed and
extensively investigated over the past decade
\cite{Herlocker2004,Konstan1997,Huang2004}. When predicting the
potential interests of a given user, such approach first identifies
a set of similar users from the past records and then makes a
prediction based on the weighted combination of those similar users'
opinions. Despite its wide applications, collaborative filtering
suffers from several major limitations including system scalability
and accuracy \cite{Sarwar2000}. Recently, some physical dynamics,
including mass diffusion \cite{Zhou2007,Zhang2007b}, heat conduction
\cite{Zhang2007a} and trust-based model \cite{Schweitzer2008}, have
found their applications in personal recommendations. These physical
approaches have been demonstrated to be of both high accuracy and
low computational complexity \cite{Zhang2007a,Zhou2007,Zhang2007b}.
However, the algorithmic accuracy and computational complexity may
be very sensitive to the statistics of data sets. For example, the
algorithm presented in Ref. \cite{Zhou2007} runs much faster than
standard CF if the number of users is much larger than that of
objects, while when the number of objects is huge, the advantage of
this algorithm vanishes because its complexity is mainly determined
by the number of objects (see Ref. \cite{Zhou2007} for details). In
order to increase the system scalability and accuracy of standard
CF, we introduce a network-based recommendation algorithm with
spreading activation, namely SA-CF. In addition, two free
parameters, $\beta$ and $N$ are presented to increase the accuracy
and personality.

\section{Method}
Denoting the object set as $O = \{o_1,o_2, \cdots, o_n\}$ and user
set as $U$ = $\{u_1, u_2,$ $\cdots,$  $u_m\}$, a recommendation
system can be fully described by an adjacent matrix $A=\{a_{ij}\}\in
R^{n,m}$, where $a_{ij}=1$ if $o_i$ is collected by $u_j$, and
$a_{ij}=0$ otherwise. For a given user, a recommendation algorithm
generates a ranking of all the objects he/she has not collected
before.

Based on the user-object matrix $A$, a user similarity network can
be constructed, where each node represents a user, and two users are
connected if and only if they have collected at least one common
object. In the standard CF, the similarity between $u_i$ and $u_j$
can be evaluated directly by a correlation function:
\begin{equation}
s_{ij}^c=\frac{\sum_{l=1}^n a_{li}a_{lj}}{\min\{k(u_i),k(u_j)\}},
\end{equation}
where $k(u_i)=\sum_{l=1}^na_{li}$ is the degree of user $u_i$.
Inspired by the diffusion process presented by Zhou {\it et al.}
\cite{Zhou2007}, we assume a certain amount of resource (e.g.
recommendation power) is associated with each user, and the weight
$s_{ij}$ represents the proportion of the resource, which $u_j$
would like to distribute to $u_i$. Following a network-based
resource-allocation process \cite{Ou2007} where each user
distributes his/her initial resource equally to all the objects
he/she has collected, and then each object sends back what it has
received to all the users collected it, the weight $s_{ij}$ (the
fraction of initial resource $u_j$ eventually gives to $u_i$) can be
expressed as:
\begin{equation}\label{equation2}
s_{ij}=\frac{1}{k(u_j)}\sum^n_{l=1}\frac{a_{li}a_{lj}}{k(o_l)},
\end{equation}
where $k(o_l)=\sum^m_{i=1}a_{li}$ denotes the degree object $o_l$.
Using the spreading process, the user correlation network can be
constructed, whose edge weight is obtained by Eq. (\ref{equation2}).
For the user-object pair $(u_i,o_j)$, if $u_i$ has not yet collected
$o_j$ (i.e. $a_{ji}=0$), the predicted score, $v_{ij}$, is given as
\begin{equation}\label{equation1}
v_{ij}=\frac{\sum_{l=1,l\neq i}^ms_{li}a_{jl}}{\sum_{l=1,l\neq
i}^ms_{li}}.
\end{equation}
From the definition of Eq.(\ref{equation1}), one can get that, to a
target user, all of his neighbors' collection information would
affect the recommendation results, which is different with the
definition reachability \cite{PRE76}. Based on the definitions of
$s_{ij}$ and $v_{ij}$, SA-CF can be given. The framework of the
algorithm is organized as follows: (I) Calculate the user similarity
matrix $\{s_{ij}\}$ based on the spreading approach; (II) For each
user $i$, obtain the score $v_{ij}$ on every object not being yet
collected by $j$; (III) Sort the uncollected objects in descending
order of $v_{ij}$, and those in the top will be recommended.

\section{Numerical results}

To test the algorithmic accuracy and personality, we use a benchmark
data-set, namely \emph{MovieLens} \cite{ex3}. The data consists of
1682 movies (objects) and 943 users, who vote movies using discrete
ratings 1-5. Hence we applied the coarse-graining method previously
used in Refs. \cite{Zhou2007,Zhou2007b}: A movie is set to be
collected by a user only if the giving rating is larger than 2. The
original data contains $10^5$ ratings, 85.25\% of which are $\geq
3$, thus the user-object (user-movie) bipartite network after the
coarse gaining contains 85250 edges. To test the recommendation
algorithms, the data set is randomly divided into two parts: the
training set contains 90\% of the data, and the remaining 10\% of
data constitutes the probe. The training set is treated as known
information, while no information in the probe set is allowed to be
used for prediction.

A recommendation algorithm should provide each user with an ordered
queue of all its uncollected objects. It should be emphasized that,
the length of queue should not be given artificially, because of the
fact that the number of uncollected movies for different users are
different. For an arbitrary user $u_i$, if the relation $u_i$-$o_j$
is in the probe set (according to the training set, $o_j$ is an
uncollected object for $u_i$), we measure the position of $o_j$ in
the ordered queue. For example, if there are $L_i=100$ uncollected
movies for $u_i$, and $o_j$ is the 10th from the top, we say the
position of $o_j$ is $10/L_i$, denoted by $r_{ij}=0.1$. Since the
probe entries are actually collected by users, a good algorithm is
expected to give high recommendations to them, thus leading to small
$r_{ij}$. Therefore, the mean value of the position $r_{ij}$,
$\langle r\rangle$ (called \emph{ranking score} \cite{Zhou2007}),
averaged over all the entries in the probe, can be used to evaluate
the algorithmic accuracy: the smaller the ranking score, the higher
the algorithmic accuracy, and vice verse. Implementing the SA-CF and
CF \cite{ex1}, the average value of ranking score are $0.12187\pm
0.02406$ and $0.13069\pm 0.0571$ \cite{ex2}. Clearly, under the
simplest initial configuration, subject to the algorithmic accuracy,
the SA-CF algorithm outperforms the standard CF.

\begin{figure}[b]
\center\scalebox{1.0}[1.0]{\includegraphics{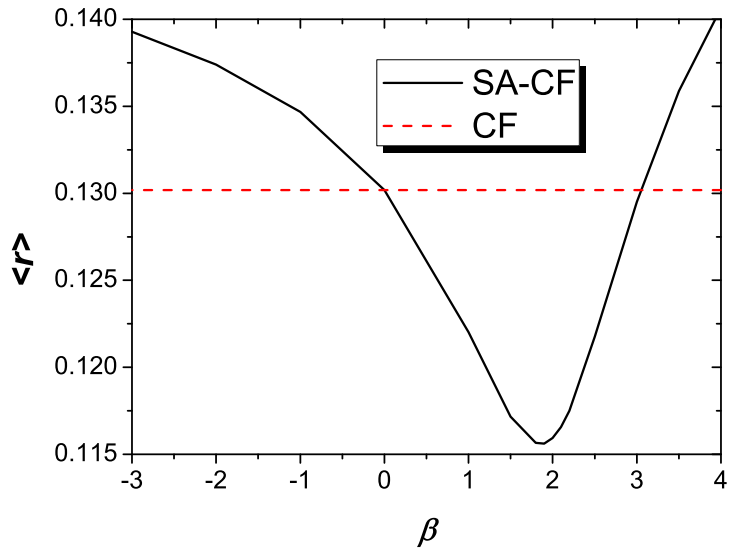}} \caption{(Color
online) $\langle r\rangle$ vs. $\beta$. The black solid and red dash
curves represent the performances of SA-CF and CF, respectively. All
the data points are obtained by averaging over ten independent runs
with different data-set divisions.}\label{Fig1.1}
\end{figure}

\begin{figure}[b]
\center\scalebox{1.0}[1.0]{\includegraphics{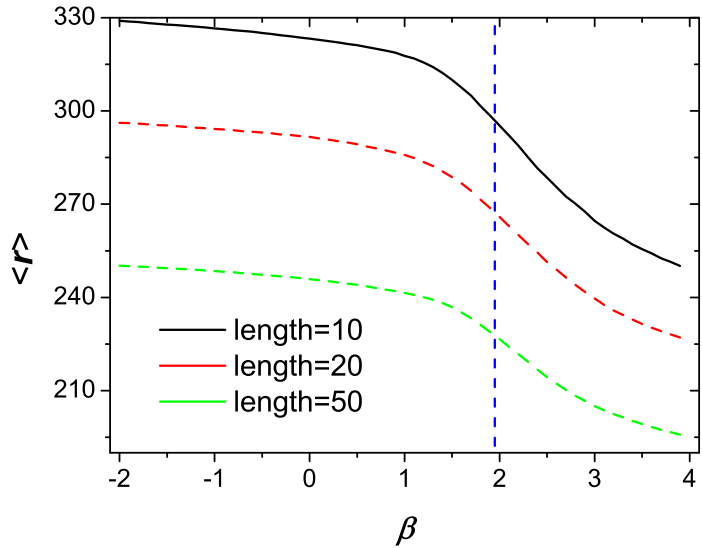}} \caption{(Color
online) The average degree of all recommended movies $\langle
k\rangle$ vs. $\beta$. The black solid, red dashed and green dotted
curves represent the cases with typical length $L=10, 20$ and 50,
respectively. The blue dot line corresponds to the optimal value
$\beta_{\rm opt}=0.19$. All the data points are obtained by
averaging over ten independent runs with different data-set
divisions.}\label{Fig1.2}
\end{figure}

\begin{figure}[b]
\center\scalebox{1.0}[1.0]{\includegraphics{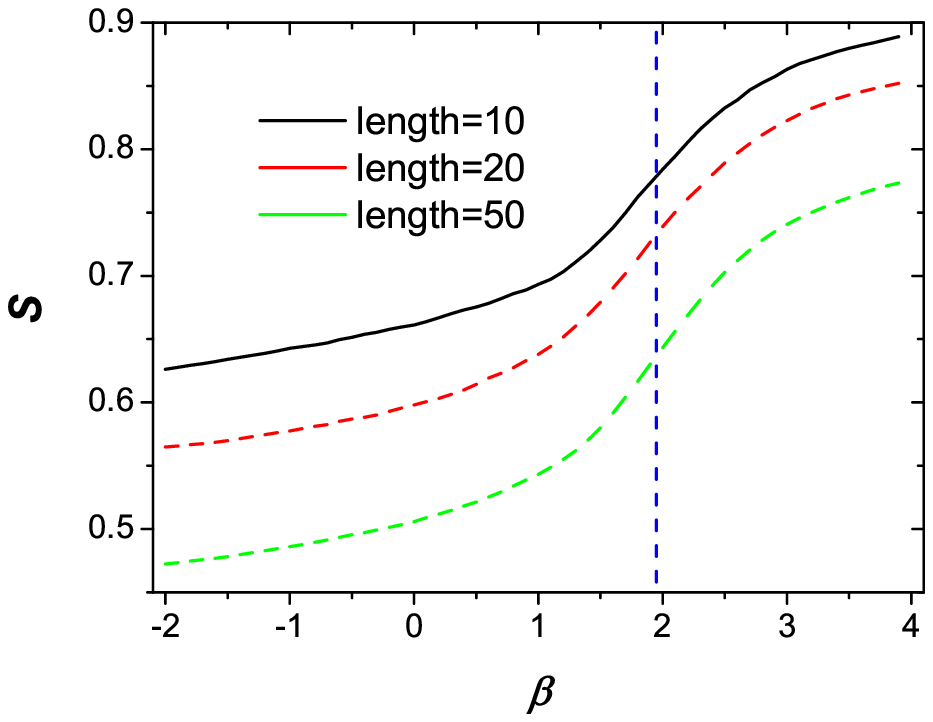}} \caption{(Color
online) $S$ vs. $\beta$. The black solid, red dashed and green
dotted curves represent the cases with typical lengths $L=10, 20$
and 50, respectively. The blue dot line corresponds to the optimal
value $\beta_{\rm opt}=0.19$. All the data points are obtained by
averaging over ten independent runs with different data-set
divisions.}\label{Fig1.3}
\end{figure}

\begin{figure}[b]
\center\scalebox{1.0}[1.0]{\includegraphics{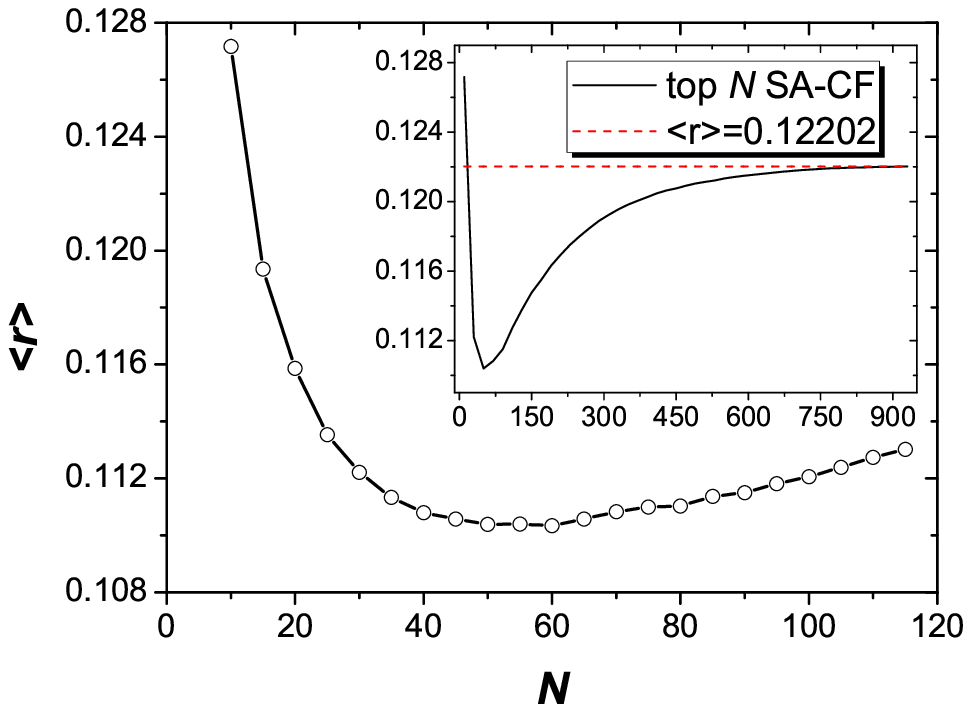}} \caption{
$\langle r\rangle$ vs. $N$. The inset shows the relation for larger
$N$. Clearly, when $N$ approaches $n$, the algorithmic accuracy is
the same as that of the SA-CF with $\langle r\rangle=0.12187\pm
0.02406$. All the data points are obtained by averaging over ten
independent runs with different data-set divisions. }\label{Fig1.4}
\end{figure}

\section{Two modified algorithms}
In order to further improve the algorithmic accuracy, we propose two
modified methods. Similar to the Ref. \cite{Zhou2007b}, taking into
account the potential role of object degree may give better
performance. Accordingly, instead of Eq. (2), we introduce a more
complicated way to get user-user correlation:
\begin{equation}
s_{ij}=\frac{1}{k(u_j)}\sum^m_{l=1}\frac{a_{li}a_{lj}}{k^{\beta}(o_l)},
\end{equation}
where $\beta$ is a tunable parameter. When $\beta=1$, this method
degenerates to the algorithm mentioned in the last section. The case
with $\beta>1$ weakens the contribution of large-degree objects to
the user-user correlation, while $\beta<1$ will enhance the
contribution of large-degree objects. According to our daily
experience, if two users $u_i$ and $u_j$ has simultaneously
collected a very popular object (with very large degree), it doesn't
mean that their interests are similar; on the contrary, if two users
both collected an unpopular object (with very small degree), it is
very likely that they share some common and particular tastes.
Therefore, we expect a larger $\beta$ (i.e. $\beta>1$) will lead to
higher accuracy than the routine case $\beta=1$.

Fig.\ref{Fig1.1} reports the algorithmic accuracy as a function of
$\beta$. The curve has a clear minimum around $\beta=1.9$, which
strongly support the above statement. Compared with the routine case
($\beta=1$), the ranking score can be further reduced by 11.2\% at
the optimal value. It is indeed a great improvement for
recommendation algorithms. Besides accuracy, the average degree of
all recommended movies $\langle k\rangle$ and the mean value of
Hamming distance $S$ \cite{Zhou2007b} are taken in account to
measure the algorithmic personality. The movies with higher degrees
are more popular than the ones with smaller degrees. The personal
recommendation should give small $\langle k\rangle$ to fit the
special tastes of different users. Fig.\ref{Fig1.2} reports the
average degree of all recommended movies as a function of $\beta$.
One can see from fig.\ref{Fig1.2} that the average degree is
negatively correlated with $\beta$, thus depressing the
recommendation power of high-degree objects gives more opportunity
to the unpopular objects. The Hamming distance, $S=\langle
H_{ij}\rangle$, is defined by the mean value among any two
recommended lists of $u_i$ and $u_j$, where $H_{ij}=1-Q/L$, $L$ is
the list length and $Q$ is the overlapped number of objects in the
two users' recommended lists. Fig.\ref{Fig1.3} shows the positively
correlation between $S$ and $\beta$, in according with the
simulation results in fig.\ref{Fig1.2}, which indicates that
depressing the influence of high-degree objects makes the
recommendations more personal. The above simulation results indicate
that SA-CF outperforms CF from the viewpoints of accuracy and
personality.

Besides the algorithmic accuracy and personality, the computational
complexity should also be taken into account. Actually, we argue
that a better algorithm should simultaneously require less
computation and generate higher accuracy. Note that, the
computational complexity of Eq. (3) is very high if the number of
user, $m$, is huge. Actually, the majority of user-user similarities
are very small, which contribute little to the final recommendation.
However, those inconsequential items, corresponding to the less
similar users, dominate the computational time of Eq. (3).
Therefore, we propose a modified algorithm, so-called top-$N$ SA-CF,
which only considers the $N$ most similar users' information to any
given user. That is to say, in the top-$N$ SA-CF, the sum in Eq. (3)
runs over only the $N$ most similar users of $u_i$. In the process
of calculation the similarity matrix $s_{ij}$, to each other, we can
simultaneously record its most similar users. When $m \gg N$, the
additional computing time for top-$N$ similar users are remarkably
shorter than what we can save from the traditional calculation of
Eq. (3). More surprisingly, as shown in Fig.\ref{Fig1.4}, with
properly chosen $N$, this algorithm not only reduces the
computation, but also enhances the algorithmic accuracy. This
property is of practical significance, especially for the huge-size
recommender systems. From figures \ref{Fig1.2} and \ref{Fig1.3}, one
can find that, to the same $\beta$ range, the anticorrelations
between $\langle k\rangle$, $S$ and $\beta$ are different in
different $\beta$ range. Maybe there is a phase transition in the
anticorrelations. Because this paper mainly focuses on the accuracy
and personality of the recommendation algorithms, this issue would
be investigated in the future.

\section{Conclusions}

In this paper, the spreading activation approach is presented to
compute the user similarity of the collaborative filtering
algorithm, named SA-CF. The basic SA-CF has obviously higher
accuracy than the standard CF. Ignoring the degree-degree
correlation in user-object relations, the algorithmic complexity of
SA-CF is $\mathbb{O}(m\langle k_u\rangle\langle k_o\rangle+mn\langle
k_o\rangle)$, where $\langle k_u\rangle$ and $\langle k_o\rangle$
denote the average degree of users and objects. Correspondingly, the
algorithmic complexity of the standard CF is $\mathbb{O}(m^2\langle
k_u\rangle+mn\langle k_o\rangle)$, where the first term accounts for
the calculation of similarity between users, and the second term
accounts for the calculation of the predictions. In reality, the
number of users, $m$, is much larger than the average object degree,
$\langle k_o\rangle$, therefore, the computational complexity of
SA-CF is much less than that of the standard CF. The SA-CF has great
potential significance in practice.

Furthermore, we proposed two modified algorithms based on SA-CF. The
first algorithm weakens the contribution of large-degree objects to
user-user correlations, and the second one eliminates the influence
of less similar users. Both the two modified algorithm can further
enhance the accuracy of SA-CF. More significantly, with properly
choice of the parameter $N$, top-$N$ SA-CF can simultaneously
reduces the computational complexity and improves the algorithmic
accuracy.

A natural question on the presented algorithms is whether these
algorithms are robust to other data-sets or random recommendation?
To SA-CF, the answer is yes, because it would get the user
similarity more accurately. While to the two modified algorithms,
the answer is no. Since both of the two modified algorithms
introduced the tunable parameters $\beta$ and $N$, the optimal
values of different data-sets are different. The further work would
focus on how to find an effective way to obtain the optimal value
exactly, then the modified algorithms could be implemented more
easily.

%\begin{acknowledgments}
We are grateful to Tao Zhou, Zolt\'{a}n Kuscsik, Yi-Cheng Zhang and
Met\'u$\breve{s}$ Medo for their greatly useful discussions and
suggestions. This work is partially supported by SBF (Switzerland)
for financial support through project C05.0148 (Physics of Risk),
the National Basic Research Program of China (973 Program No.
2006CB705500), the National Natural Science Foundation of China
(Grant Nos. 60744003, 10635040, 10532060, 10472116), GQ was
supported by the Liaoning Education Department (Grant No. 20060140).
%\end{acknowledgments}


\begin{thebibliography} {1}
\bibitem{ex1} The predicted results of standard CF can be directly
obtained by applying Eq. (3) using $s_{li}^c$ instead of $s_{li}$.
\end{thebibliography}
\end{document}